\documentclass[journal]{IEEEtran}

\ifCLASSINFOpdf
\else
   \usepackage[dvips]{graphicx}
\fi
\usepackage{url}

\usepackage{graphicx}

\usepackage{enumitem}
\usepackage{booktabs} 
\usepackage{caption} 
\usepackage{adjustbox} 
\usepackage{array} 
\usepackage{multirow}
\usepackage[table,xcdraw]{xcolor}
\usepackage{svg}
\usepackage{float}
\usepackage[most]{tcolorbox}
\usepackage{pgfplots}  
\usepackage{tikz}
\usepackage{textcomp}
\usepackage{cite}
\usepackage{algorithmic}
\usepackage{hyperref}
\hypersetup{hypertex=true,
            colorlinks=true,
            linkcolor=green,
            anchorcolor=green,
            citecolor=red}
\usepackage{amsmath,amssymb,amsfonts}

\def\BibTeX{{\rm B\kern-.05em{\sc i\kern-.025em b}\kern-.08em
    T\kern-.1667em\lower.7ex\hbox{E}\kern-.125emX}}

\begin{document}

\title{Context-Aware Weakly Supervised Image Manipulation Localization with SAM Refinement}

\author{
        Xinghao Wang, 
        Tao Gong\textsuperscript{\textasteriskcentered},
        Qi Chu,
        Bin Liu and
        Nenghai Yu
    
\thanks{$^*$ Corresponding author: tgong@ustc.edu.cn.}
\thanks{The authors are with the University of Science and Technology of China, HeFei 230026, China. (e-mail: \{wxhwxhwxh\}@mail.ustc.edu.cn,
            \{tgong, qchu, flowice, ynh\}@ustc.edu.cn).}
}

\markboth{}
{Shell \MakeLowercase{\textit{et al.}}: Bare Demo of IEEEtran.cls for IEEE Journals}
\maketitle

\begin{abstract}
Malicious image manipulation poses societal risks, increasing the importance of effective image manipulation detection methods. Recent approaches in image manipulation detection have largely been driven by fully supervised approaches, which require labor-intensive pixel-level annotations. Thus, it is essential to explore weakly supervised image manipulation localization methods that only require image-level binary labels for training. However, existing weakly supervised image manipulation methods overlook the importance of edge information for accurate localization, leading to suboptimal localization performance. To address this, we propose a Context-Aware Boundary Localization (CABL) module to aggregate boundary features and learn context-inconsistency for localizing manipulated areas. Furthermore, by leveraging Class Activation Mapping (CAM) and Segment Anything Model (SAM), we introduce the CAM-Guided SAM Refinement (CGSR) module to generate more accurate manipulation localization maps. By integrating two modules, we present a novel weakly supervised framework based on a dual-branch Transformer-CNN architecture. Our method achieves outstanding localization performance across multiple datasets.
\end{abstract}

\begin{IEEEkeywords}
Image Manipulation Localization, Weakly Supervised, Context-Aware Boundary Localization
\end{IEEEkeywords}

\IEEEpeerreviewmaketitle

\section{Introduction}

\IEEEPARstart{W}{ith} the proliferation of maliciously forged images, the localization of manipulated regions has become critical for combating misinformation. Currently, most image manipulation localization methods leverage both image-level binary classification labels and pixel-level mask annotations for fully supervised training\cite{b19,b48}. Early approaches leveraged noise, boundaries, and color inconsistencies \cite{b41,b42}, while recent methods focus on content-agnostic features: CAT-Net \cite{b1} learns compression artifacts; MVSS-Net \cite{b2} uses multi-scale noise; Mantra-Net \cite{b3} detects operational boundaries; and PROMPT-IML \cite{b4} aligns semantic and high-frequency features via pre-trained models.

\begin{figure*}[htbp]
\centering
\setlength{\belowcaptionskip}{-0.2cm}   
\includegraphics[width=0.8\textwidth]{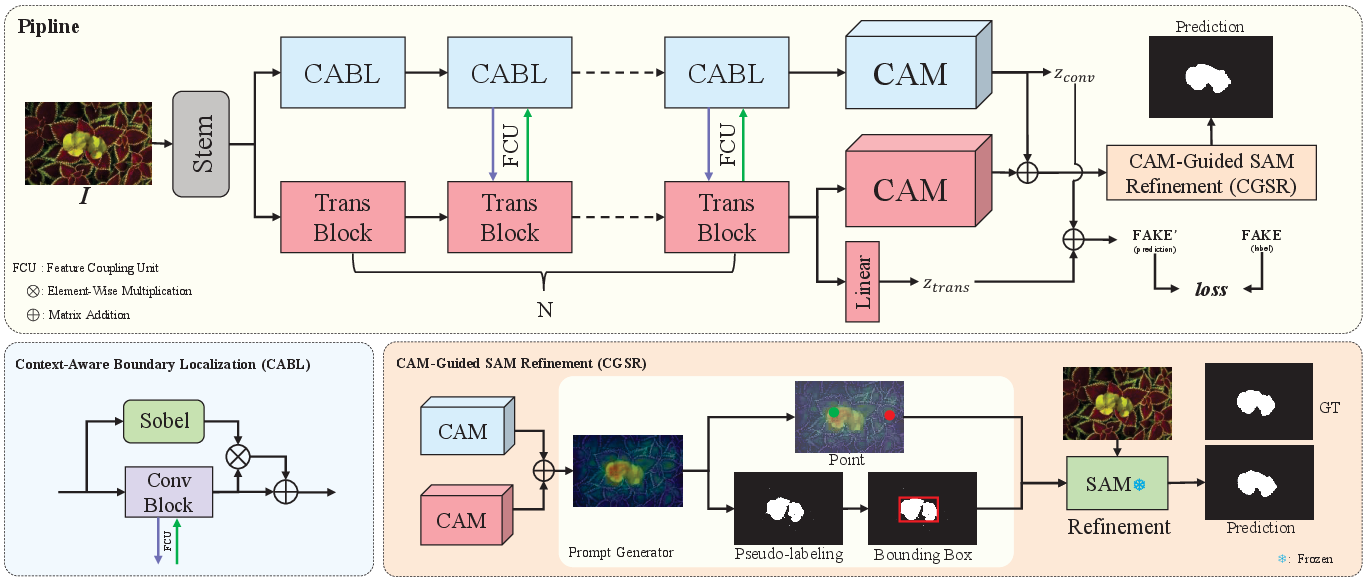}
\captionsetup{font=small}
\caption{The framework of our method: We integrate our custom-designed CABL module and CGSR module into the backbone. We train the model using only images and image-level binary labels. Sobel\cite{b24} denotes the Sobel operator, Stem generates patch embeddings, ConvBlock and TransBlock\cite{b8} are composed of convolution layers and transformer\cite{b30} layers respectively. The green point in the Prompt Generator indicates the positive point prompt, the red point represents the negative point prompt, and the red box denotes the bounding box prompt.}
\label{fig:pipline}
\vspace{-0.4cm}  
\end{figure*}

\begin{figure}[t]
\centering
\setlength{\belowcaptionskip}{-0.3cm}   
\includegraphics[width=0.3\textwidth]{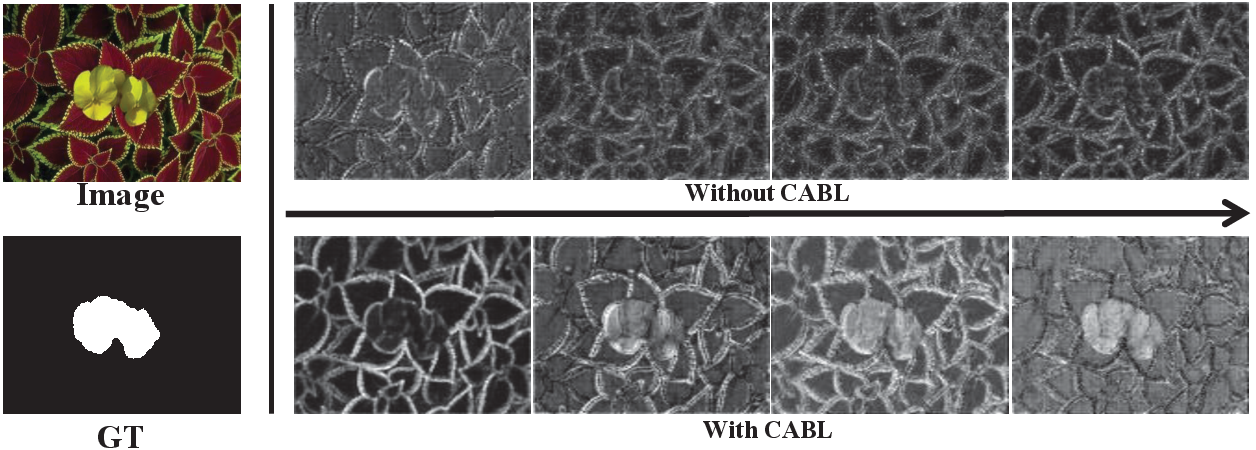}
\captionsetup{font=small}
\caption{Right of vertical line: first row (w/o CABL) shows feature maps of first 4 blocks; second row (w/ CABL) shows enhanced maps. Arrows indicate deeper block directions.}
\vspace{-0.1cm}  
\label{fig:CABL}
\end{figure}

Despite significant advances in fully supervised image manipulation localization methods\cite{b19,b49}, the growing diversity and complexity of manipulation techniques have made acquiring comprehensive pixel-level mask annotations increasingly impractical. Dense annotation of manipulated regions is both challenging and time-consuming. As a result, recent research has shifted towards weakly supervised approaches to enhance generalization and localization performance. For instance, WSCL\cite{b5} introduces multi-source consistency and inter-patch consistency modules by exploiting the self-consistency properties of manipulated images. While effective for manipulation detection tasks, its localization performance is limited by inadequate consideration of boundary information, which is crucial for accurate manipulation localization.

Image manipulation techniques like copy-move\cite{b34,b36} and splicing\cite{b33} involve non-generative operations in which manipulated regions are copied from authentic areas. As a result, the noise and texture within manipulated regions often resemble those of real regions, making the boundaries of manipulated areas crucial for localization. To better learn boundary information, we propose a Context-Aware Boundary Localization (CABL) module that utilizes context-inconsistency at the boundaries to more precisely identify manipulated regions. Furthermore, previous methods\cite{b5,b47} in weakly supervised image manipulation localization have largely overlooked the potential of Class Activation Mapping (CAM)\cite{b13}. In our approach, we compute CAM maps using the output of a dual-branch Transformer-CNN network to generate coarse masks of manipulated regions. To refine these coarse masks, we leverage the strong capabilities of the Segment Anything Model (SAM)\cite{b23} through a CAM-Guided SAM Refinement (CGSR) module. Specifically, we use the bounding rectangle of the largest connected component from the coarse masks and the positive and negative sampling points from the CAM\cite{b13} without maximum-minimum normalization as visual prompts for SAM\cite{b23}. SAM\cite{b23} can generate more accurate localization masks using these visual prompts, without requiring additional training. By integrating these two modules, we propose a novel framework that combines CABL and CGSR within a dual-branch Transformer-CNN architecture. This framework can learn the context-inconsistency of the manipulated images under weakly supervised training conditions, enabling precise localization of the manipulated regions.

Experiments validate that our method achieves state-of-the-art performance in both manipulation detection and localization, effectively identifying and segmenting manipulated regions across diverse scenarios. The principal contributions of this work include:
\begin{enumerate}[label=(\arabic*)]
  \item We propose the \textbf{Context-Aware Boundary Localization (CABL) module} that explicitly learns context-inconsistency at manipulated boundaries, enabling precise localization of manipulated regions.
  \item We propose the \textbf{CAM-Guided SAM Refinement (CGSR) module}, employing CAM\cite{b13} to generate the coarse-grained masks and achieving more precise localization results base on SAM's\cite{b23} powerful segmentation capabilities without additional training.
  \item  We train the model using only image-level binary classification labels from the CASIAv2\cite{b6} dataset and evaluate its performance on four unseen test datasets, demonstrating its effectiveness.
\end{enumerate}

\section{METHODOLOGY}
\label{sec:meth0dology}

\subsection{Overview}
\label{ssec:overview}

\begin{table*}[ht]
\centering
\setlength{\belowcaptionskip}{-0.3cm}   
\caption{Comparison of Image Manipulation Detection and Localization Methods across Different Datasets$^{\mathrm{1},\mathrm{2}}$}
\adjustbox{max width=0.85\textwidth}{
\begin{tabular}{l|l|cc|cc|cc|cc|c|c}
\bottomrule
\multirow{2}{*}{} & \multirow{2}{*}{\textbf{Method}} & \multicolumn{2}{c|}{\textbf{CASIAv1}} & \multicolumn{2}{c|}{\textbf{Columbia}} & \multicolumn{2}{c|}{\textbf{Coverage}} & \multicolumn{2}{c|}{\textbf{IMD2020}} & \multicolumn{2}{c}{\textbf{Average}} \\ 
 &  & I-AUC & P-F1 & I-AUC & P-F1 & I-AUC & P-F1 & I-AUC & P-F1 & I-AUC & P-F1  \\ 
\hline

\cellcolor{red!10} & NOII \cite{b14} & 0.500  &0.157 & 0.500 &0.311& 0.500 &0.205& 0.500 & 0.124& 0.500 & 0.199 \\  
 \multirow{-2}{*}{\rotatebox{90}{\textbf{Un.$^{\mathrm{3}}$}}}
 \cellcolor{red!10}& CFAl \cite{b15} & 0.482  &0.140 & 0.344 &0.320& 0.525&0.188& 0.500 & 0.111 & 0.500 & 0.190 \\ 
\hline

  \cellcolor{green!10}& Mantra-Net \cite{b16} & 0.141 &0.155& 0.701 &0.364& 0.490  &0.286& 0.719 &0.122 & 0.513 & 0.231 \\  
  \cellcolor{green!10}& CAT-Net \cite{b1} & 0.630&0.276&  0.782 &0.352& 0.572 &0.134& 0.721 & 0.102 & 0.693 & 0.216 \\ 
  \cellcolor{green!10}& MVSS-Net \cite{b2} & 0.937&0.452&  0.980 &0.638& 0.731 &0.453& 0.656 & 0.260 & 0.826 & 0.451 \\
  \cellcolor{green!10}& GSR-Net \cite{b18} & 0.502&0.387&  0.502&0.613& 0.515&0.285& 0.505& 0.175 & 0.506 & 0.365 \\
  \multirow{-5}{*}{\rotatebox{90}{\textbf{Full.}}}
  \cellcolor{green!10} & FCN+DA \cite{b19} & 0.796  &0.441& 0.762 &0.223& 0.541 &0.199& 0.746 &0.270& 0.711 &0.283\\

\hline

\cellcolor{brown!20}& MIL-FCN \cite{b20} & 0.647 & 0.117& 0.807 & 0.089& 0.542 &0.121& 0.578 & 0.097& 0.644 &0.106\\  
\cellcolor{brown!20}& MIL-FCN\cite{b20}+WSCL\cite{b5} & \textbf{0.829}  &\underline{0.172}& \underline{0.920}& 0.270& 0.584  &0.178& \underline{0.733}  &\underline{0.193}& \underline{0.766} &0.203 \\  
\cellcolor{brown!20}& Araslanov and Roth \cite{b21} & 0.642 &0.112 & 0.773 & 0.102& 0.560 & 0.127& 0.665 &0.094& 0.660 & 0.109 \\  
\cellcolor{brown!20}& Araslanov and Roth \cite{b21}+WSCL\cite{b5} & 0.796  &0.153& 0.917 & \underline{0.362}& \underline{0.591} &\textbf{0.201}& 0.701  &0.173& 0.751&\underline{0.222}\\  
\rowcolor{gray!30}
\multirow{-5}{*}{\rotatebox{90}{\textbf{Weak.}}}
 \cellcolor{brown!20}  & \textbf{Ours} & \underline{0.816} &\textbf{0.300}& \textbf{0.963} & \textbf{0.382}& \textbf{0.658} &\underline{0.190}& \textbf{0.783} &\textbf{0.368}& \textbf{0.805} & \textbf{0.310} \\

\toprule
\multicolumn{12}{l}{$^{\mathrm{1}}$The best and second best results in the weakly supervised method are indicated in \textbf{boldface} and \underline{underlined}, respectively.}\\
\multicolumn{12}{l}{$^{\mathrm{2}}$All localization results in \textbf{our} method were refined using CGSR.}\\
\multicolumn{12}{l}{$^{\mathrm{3}}$Un. means unsupervised methods; Full. means fully supervised methods; Weak. means weakly supervised methods.}
\end{tabular}}
\vspace{-0.4cm}  
\label{tab:1}
\end{table*}

\begin{table}[t]
\centering
\captionsetup{font=small}
\caption{Ablation Study of \textbf{CABL} and \textbf{CGSR} Modules on Coverage and IMD2020 Datasets}
\resizebox{0.4\textwidth}{!}{
\begin{tabular}{cc|cc|cc|c|c}
\bottomrule
\multirow{2}{*}{\textbf{CABL}}&\multirow{2}{*}{\textbf{CGSR}} & \multicolumn{2}{c|}{\textbf{Coverage}} & \multicolumn{2}{c|}{\textbf{IMD2020}} & \multicolumn{2}{c}{\textbf{Avg}} \\ 
& & I-AUC & P-F1 & I-AUC &P-F1 & I-AUC & P-F1 \\ \hline
- & - & 0.649 & 0.154 & 0.767 & 0.342 & 0.708 & 0.248 \\ 
\checkmark & - & 0.658& 0.177 & 0.783 & 0.361 & 0.721 & 0.269\\ 
- & \checkmark & 0.649& 0.175 & 0.767 & 0.353 & 0.708 & 0.264\\ 
\rowcolor{gray!40}
\checkmark & \checkmark & 0.658 & 0.190 & 0.783 & 0.368 & \textbf{0.721} &  \textbf{0.279}\\ 
\toprule
\end{tabular}}
\vspace{-0.1cm}  
\label{tab:a1}
\end{table}

\begin{table}[t]
\centering
\setlength{\belowcaptionskip}{-0.1cm}   
\captionsetup{font=small}
\caption{Ablation Study of \textbf{Different CABL Structures} on Coverage and IMD2020 Datasets$^{\mathrm{1}}$}
\resizebox{0.4\textwidth}{!}{
\begin{tabular}{c|cc|cc|c|c}
\bottomrule
\multirow{2}{*}{\textbf{Structure}} & \multicolumn{2}{c|}{\textbf{Coverage}} & \multicolumn{2}{c|}{\textbf{IMD2020}} & \multicolumn{2}{c}{\textbf{Avg}} \\ 
 & I-AUC &P-F1 & I-AUC &  P-F1 & I-AUC & P-F1 \\ \hline
\textsc{i} & 0.617 & 0.130 & 0.774 & 0.257 & 0.696 & 0.194 \\ 
\textsc{ii} & 0.623 & 0.154 & 0.767 & 0.334 & 0.695 & 0.244 \\ 
\rowcolor{gray!30}
\textbf{\textsc{iii}} & 0.658 & 0.177 & 0.783 & 0.361 & \textbf{0.721} & \textbf{0.269}\\ 
\toprule
    \multicolumn{7}{l}{$^{\mathrm{1}}$All localization results were obtained without CGSR.}
\end{tabular}}
\vspace{-0.6cm}  
\label{tab:a2}
\end{table}

As shown in Fig.\ref{fig:pipline}, the framework first processes the input image through a Stem module to extract hierarchical features, followed by two parallel branches: N-block CABL module explicitly modeling boundary context-inconsistency for manipulation localization, and TransBlock \cite{b8} capturing global semantic features via self-attention. The two branches use the Feature Coupling Unit (FCU)\cite{b11} to integrate local and global features. Classification is supervised with Binary CrossEntropy Loss (BCE)\cite{b31}. The final layer of each branch undergoes global average pooling, and the results are weighted to generate the Class Activation Map (CAM)\cite{b13} for each class (Sec.\ref{ssec:CABL}). These coarse-grained masks are then used to create prompts. The prompts are subsequently input into the Segment Anything Model (SAM)\cite{b23} for refinement, yielding final localization predictions without additional training (Sec.\ref{ssec:cgsr}).

\subsection{Context-Aware Boundary Localization (CABL)}
\label{ssec:CABL}

\begin{figure}[t]
\centering
\setlength{\belowcaptionskip}{-0.1cm}   
\includegraphics[width=0.3\textwidth]{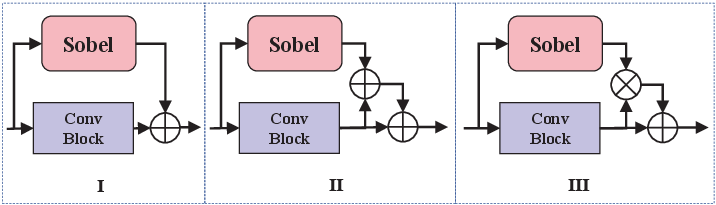}
\captionsetup{font=small}
\caption{Originally designed for three possible CABL structures, structure III was eventually used as CABL}
\vspace{-0.2cm}  
\label{fig:different CABL}
\end{figure}

\begin{figure}[t]
    \centering
    \setlength{\belowcaptionskip}{-0.1cm}   
    \includegraphics[width=0.35\textwidth]{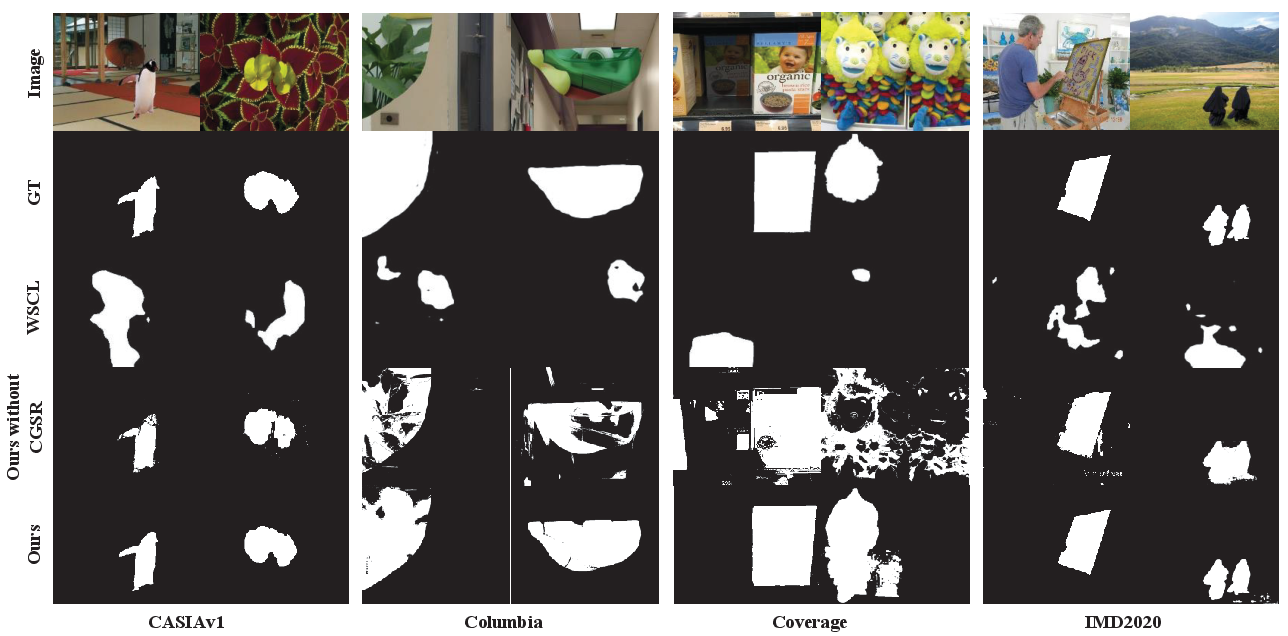}
    \captionsetup{font=small}
    \caption{Qualitative results on four datasets. From top to bottom: Image, GT, WSCL+MIL-FCN, Ours w/o and w/ SAM\cite{b23} refinement.}
    \label{fig:qualitative_results}
    \vspace{-0.4cm}  
\end{figure}

Image manipulation tasks such as copy-move \cite{b34,b36} and splicing \cite{b33} typically involve regions sourced from genuine images where internal textures and noise are similar to real regions, the localization model must focus on boundary areas between real and manipulated regions. However, weakly supervised methods face challenges in this context, as they rely solely on image-level binary labels and cannot utilize pixel-level masks, complicating direct boundary learning. To address this, we propose utilizing non-learnable edge detection operators to extract high-frequency boundary information, which enhances the model's focus on edge features.

The Context-Aware Boundary Localization (CABL) module is designed to capture context-inconsistent boundary information within feature maps. At each block, CABL extracts contextual features, applies an edge detection operator to amplify high-frequency boundary signals, and performs element-wise multiplication and residual connections \cite{b39} between boundary and image features. This aggregated boundary information is propagated through subsequent blocks. We adopt the Sobel operator \cite{b24} as the edge detector due to its computational efficiency and noise suppression capability. While the Prewitt operator \cite{b24} shares a similar design, it underperforms in noise robustness (Sec.\ref{sssec:edge}). The CABL structure is illustrated in Fig.\ref{fig:CABL}, and its feature computation is formalized as:

\begin{equation}
F_i=S(F_{i-1}) \otimes C(F_{i-1}) \oplus C(F_{i-1})
\label{eqa:cabl}
\end{equation}
where \( F_i \) represents the features at the \( i \)-th block, \( S(\cdot) \) extracts boundary features via Sobel filtering\cite{b24}, \( C(\cdot) \)represents convolutional features from the prior block, \( \otimes \) indicates element-wise multiplication, and  \( \oplus \) denotes matrix addition.

\subsection{CAM-Guided SAM Refinement (CGSR)}
\label{ssec:cgsr}

Class Activation Mapping (CAM) \cite{b13} is a widely used technique in weakly supervised learning that visualizes regions of an image most influential for classification. Our method enhances CAM maps by fusing convolutional feature maps from the CABL branch with attention weights from the TransBlock \cite{b8} (Fig.\ref{fig:pipline}), followed by a weighted summation \cite{b25}. This combined feature representation is further integrated with the classification head weights of the TransBlock to produce higher-quality CAMs, which serve as coarse-grained masks. The CAM generation process is formalized as:

{\small
\begin{equation}
\text{CAM}=\text{CAM}_{\text{Trans}} \cup \left[ \frac{1}{L} \sum_{l} \text{softmax}\left( \frac{Q^l K^{l\mathrm{T}}}{\sqrt{D/S}} \right) \otimes \text{CAM}_{\text{Conv}} \right]
\label{eqa:cam}
\end{equation}
}
where $\mathrm{CAM_{Trans}}$ denotes the CAM\cite{b13} generated by the TransBlock, and $\mathrm{CAM_{Conv}}$ represents the CAM \cite{b13} from the ConvBlock. The symbol $\cup$ represents the union of the two CAM maps, and $\otimes$ indicates element-wise multiplication. $Q$ and $K$ are query and key matrices from the attention mechanism. \( D \) is the feature dimension. \( L \) is the total number of blocks, and \( l \) indexes individual blocks.

However, directly employing these coarse-grained masks as prediction masks may result in incomplete and imprecise localization of the manipulated regions. To address this, we incorporate the Segment Anything Model (SAM)\cite{b23}, a robust segmentation model that requires no additional training and relies only on the original image and prompts to generate accurate segmentation results. As shown in the prompt generator of CGSR in Fig.\ref{fig:pipline}, We use the bounding rectangle of the largest connected component in the coarse-grained mask as the bounding box prompt. Additionally, we select one maximum value point (green point in Fig.\ref{fig:pipline}) and one minimum value point (red point in Fig.\ref{fig:pipline}) from the Class Activation Map (CAM)\cite{b13} without maximum-minimum normalization, corresponding to the highest and lowest manipulation probabilities. These two points serve as positive and negative sample point prompts. These prompts are input into SAM to guide precise segmentation. The bounding box and positive point prompts delineate foreground regions, while the negative point suppresses background noise. This combination enables SAM to output high-precision localization maps without further training, effectively resolving boundary ambiguity in the initial CAM predictions.

\begin{table}[t]
\centering
\captionsetup{font=small}
\caption{Ablation Study of \textbf{CABL in Different Conv Blocks} on Coverage and IMD2020 Datasets$^{\mathrm{1}}$}
\resizebox{0.4\textwidth}{!}{
\begin{tabular}{c|cc|cc|c|c}
\bottomrule
\multirow{2}{*}{\textbf{Blocks}} & \multicolumn{2}{c|}{\textbf{Coverage}} & \multicolumn{2}{c|}{\textbf{IMD2020}} & \multicolumn{2}{c}{\textbf{Avg}} \\ 
 & I-AUC & P-F1 & I-AUC & P-F1 & I-AUC & P-F1 \\ \hline
1-4 & 0.622 & 0.141 & 0.727  & 0.258 & 0.675  & 0.200 \\ 
1-8 & 0.665 & 0.125 & 0.771  & 0.357 & 0.718 & 0.241 \\ 
\rowcolor{gray!30}
\textbf{1-12} & 0.658 & 0.177 & 0.783 & 0.361 & \textbf{0.721}  & \textbf{0.269} \\ 
\toprule
\multicolumn{7}{l}{$^{\mathrm{1}}$All localization results were obtained without CGSR.}
\end{tabular}}
\vspace{-0.2cm}  
\label{tab:a3}
\end{table}

\begin{figure}[t]
\centering
\begin{minipage}{0.4\linewidth}  
    \centering
    \begin{tikzpicture}[scale=0.4]
    \begin{axis}[
        xlabel={JPEG Compression Quality},
        ylabel={Pixel-level Localization F1},
        xlabel style={yshift=5pt},
        ylabel style={yshift=-10pt},
        xmin=50, xmax=100,
        ymin=0, ymax=0.33,
        xtick={100,90,80,70,60,50},
        grid=major,
        legend style={at={(0.35,0.98)},draw=white, anchor=north, legend columns=2, font=\small}
    ]
    \addplot[mark=triangle, green, thick] plot coordinates {(50,0.2301) (60,0.2393) (70,0.2316) (80,0.2323) (90,0.2343) (100,0.2533)};
    \addlegendentry{Baseline}
    \addplot[mark=*, orange, thick] plot coordinates {(50,0.2405) (60,0.2398) (70,0.2391) (80,0.2374) (90,0.2535) (100,0.3)};
    \addlegendentry{Ours}
    \addplot[mark=o, blue, thick] plot coordinates {(50,0.03161) (60,0.03632) (70,0.05482) (80,0.054203) (90,0.071688) (100,0.1126)};
    \addlegendentry{WICL+MIL-FCN}
    \end{axis}
    \end{tikzpicture}
\end{minipage}%
\begin{minipage}{0.4\linewidth}  
    \centering
    \begin{tikzpicture}[scale=0.4]
    \begin{axis}[
        xlabel={Gaussian Blur Kernel Size},
        ylabel={Pixel-level Localization F1},
        xlabel style={yshift=5pt},
        ylabel style={yshift=-10pt},
        xmin=0, xmax=29,
        ymin=0, ymax=0.33,
        xtick={29,23,17,11,5,0},
        grid=major,
        legend style={at={(0.65,0.98)},draw=white,anchor=north, legend columns=2, font=\small}
    ]
    \addplot[mark=triangle, green, thick] plot coordinates {(0,0.2533) (5,0.2478) (11,0.2023) (17,0.1703) (23,0.1696) (29,0.1565)};
    \addlegendentry{Baseline}
    \addplot[mark=*, orange, thick] plot coordinates {(0,0.3) (5,0.2682) (11,0.2427) (17,0.2235) (23,0.2222) (29,0.2069)};
    \addlegendentry{Ours}
    \addplot[mark=o, blue, thick] plot coordinates {(0,0.1126) (5,0.0804) (11,0.05558) (17,0.04723) (23,0.0383844) (29,0.038215)};
    \addlegendentry{\small WICL+MIL-FCN}
    \end{axis}
    \end{tikzpicture}
\end{minipage}
\caption{Robustness Evaluation of \textbf{JPEG Compression} and \textbf{Gaussian Blur} on CASIAv1 Dataset.}
\vspace{-0.4cm}  
\label{fig:rob}
\end{figure}
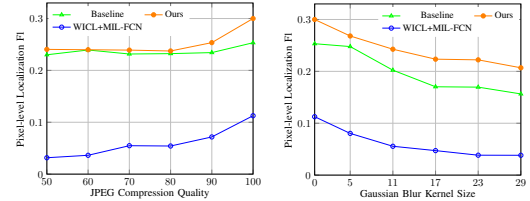

\subsection{Loss}
\label{ssec:loss}

Our weakly supervised method is trained using only image-level binary classification labels, with the loss function being the sum of the classification losses from the two network branches. The training loss of the model is Eq.~\eqref{eqa:loss}:

\begin{equation}
loss=loss_{CABL}+ loss_{Trans}
\label{eqa:loss}
\end{equation}
where \( {loss}_{CABL} \) and \( {loss}_{Trans} \) denote the Binary CrossEntropy Losses (BCE)\cite{b31} for the CABL and Trans branches, respectively.

\section{EXPERIMENT} 
\label{sec:EXPERIMENT}
\subsection{Implementation details}

We trained our model on CASIAv2\cite{b6} using only original images and image-level binary labels, without any pixel-level ground truth. For evaluation, we use CASIAv1\cite{b6}, Columbia\cite{b8}, COVERAGE\cite{b9}, and IMD2020\cite{b10} to assess both classification and localization. For image-level detection, we report the I-AUC score\cite{b19}, and for pixel-level localization, we calculate the P-F1 score, following previous methods\cite{b5}. 

The backbone we used is Conformer-S\cite{b11}, pretrained on ImageNet\cite{b22}. The model is trained for 30 epochs with a batch size of 8 using the AdamW optimizer\cite{b26}, a learning rate of 5e-5, and a weight decay of 5e-4. To address the challenges of weakly supervised training (no pixel-level annotations), we applied data augmentation techniques such as random scaling. During inference, multi-scale inputs (256×256, 512×512, and 768×768) were fed into the model, and the resulting predicted maps were aggregated to enhance final localization precision.

\subsection{Comparison with the state-of-the-art methods and Visualization}

The Table.\ref{tab:1} shows that, for image-level manipulation detection, our method achieves the highest average I-AUC among weakly supervised approaches and even outperforms some partially supervised methods. For pixel-level manipulation localization, our method also leads weakly supervised approaches in average P-F1, demonstrating excellent localization performance. While it may not outperform all fully supervised methods, its performance remains comparable to traditional fully supervised techniques. Fig.\ref{fig:qualitative_results} provides visualization examples that illustrate our method's high accuracy in segmentation and localization. As shown in the fourth row of Fig.\ref{fig:qualitative_results}, Our method produces clearer boundaries for the localized manipulated regions compared to the previous weakly supervised method (third row of Fig.\ref{fig:qualitative_results}), with a more accurate placement of the manipulated areas.

\subsection{Comparison of Robustness with Weakly Supervised Methods}

Following established conventions for evaluating weakly supervised localization robustness \cite{b5}, we assessed the P-F1 score of manipulated images subjected to JPEG compression with varying compression rates and Gaussian blur with different kernel sizes on CASIAv1\cite{b6}. As shown in Fig.\ref{fig:rob}, our localization method outperforms previous weakly supervised methods\cite{b5} in terms of robustness. 

\subsection{Ablation Study}

\subsubsection{\textbf{Impact of CABL and CGSR Modules on Model Performance}}

Ablation experiments on COVERAGE \cite{b9} and IMD2020 \cite{b10} evaluated the impact of CABL and CGSR modules. Results in Table.\ref{tab:a1} show that CABL alone improves average I-AUC by 0.013 and P-F1 by 0.021 over baseline, while CGSR adds an extra +0.008 to P-F1. The combined modules achieve the highest performance (I-AUC: +0.013, P-F1: +0.031 vs. baseline), demonstrating CABL's role in enhancing region localization accuracy and CGSR's boundary refinement capability through adaptive prediction optimization.

\subsubsection{\textbf{Ablation Study on Different CABL Structures}}

Ablation experiments on COVERAGE \cite{b9} and IMD2020 \cite{b10} compared three CABL designs (Fig.\ref{fig:different CABL}) in Table.\ref{tab:a2}. Structure III achieved the highest average classification accuracy and localization precision across both datasets, thus selected as the final configuration due to its superior performance.

\subsubsection{\textbf{Ablation Study on CABL Applied to Different Convolutional Blocks}}

\begin{table}[t]
\centering
\setlength{\belowcaptionskip}{-0.1cm}   
\captionsetup{font=small}
\caption{Ablation Study of \textbf{Different Edge Detection Operators in CABL} on Coverage and IMD2020 Datasets$^{\mathrm{1}}$}
\resizebox{0.49\textwidth}{!}{
\begin{tabular}{c|cc|cc|cc|cc|c|c}
\bottomrule
\multirow{2}{*}{\textbf{Operator}} & \multicolumn{2}{c|}{\textbf{CASIAv1}} &\multicolumn{2}{c|}{\textbf{Columbia}} & \multicolumn{2}{c|}{\textbf{Coverage}} & \multicolumn{2}{c|}{\textbf{IMD2020}} & \multicolumn{2}{c}{\textbf{Avg}} \\ 
 & I-AUC &P-F1 & I-AUC &P-F1 & I-AUC &P-F1 & I-AUC &  P-F1 & I-AUC & P-F1 \\ \hline
Prewitt & 0.824 & 0.298 & 0.971 & 0.113 & 0.654 & 0.173&0.761&0.419 &0.803&0.251\\ 
\rowcolor{gray!30}
\textbf{Sobel} & 0.816 & 0.280 & 0.963 & 0.302 & 0.658 & 0.177&0.783&0.361 &0.805&0.280\\ 
\toprule
    \multicolumn{10}{l}{$^{\mathrm{1}}$All localization results were obtained without CGSR.}
\end{tabular}}
\vspace{-0.2cm}  
\label{tab:a5}
\end{table}

\begin{table}[t]
\centering
\setlength{\belowcaptionskip}{-0.1cm}   
\captionsetup{font=small}
\caption{Ablation Study of \textbf{Different CGSR Prompt}}
\resizebox{0.45\textwidth}{!}{
\begin{tabular}{c|c|c|c|c|c}
\bottomrule
\multirow{2}{*}{\textbf{Prompt}} & \multicolumn{1}{c|}{\textbf{CASIAv1}} & \multicolumn{1}{c|}{\textbf{Columbia}} & \multicolumn{1}{c|}{\textbf{Coverage}} & \multicolumn{1}{c|}{\textbf{IMD2020}} & \multicolumn{1}{c}{\textbf{Avg}} \\ 
 & P-F1 &  P-F1 &  P-F1 & P-F1 &  P-F1 \\ \hline
Null$^{\mathrm{1}}$ & 0.280 & 0.302 & 0.177 & 0.361 & 0.280 \\
Point$^{\mathrm{2}}$ & 0.288 & 0.339 & 0.176 & 0.357 & 0.290\\ 
Box$^{\mathrm{3}}$ & 0.298 & 0.359 & 0.179 & 0.366 & 0.301 \\ 
\rowcolor{gray!30}
\textbf{Box+Point} & 0.300 & 0.382 & 0.190 & 0.368 & \textbf{0.310} \\ 
\toprule
\multicolumn{6}{l}{$^{\mathrm{1}}$Null means no CGSR;}\\
\multicolumn{6}{l}{$^{\mathrm{2}}$Point selects the highest probability point from the non-binarized CAM; }\\
\multicolumn{6}{l}{$^{\mathrm{3}}$Box selects the largest connected regions in the coarse-grained masks.}
\end{tabular}}
\vspace{-0.4cm}  
\label{tab:a4}
\end{table}

Ablation studies in Table.\ref{tab:a3} compared CABL module application to the first 4, first 8, or all CNN blocks. Applying CABL to all blocks improved average I-AUC by 0.046 (vs. first 4 blocks) and 0.003 (vs. first 8 blocks), with P-F1 gains of 0.069 and 0.028, respectively. This suggests that shallow CABL layers extract texture features, while deeper layers capture high-frequency edge information \cite{b2}, justifying the full-block implementation.

\subsubsection{\textbf{Ablation Study on Different Edge Detection Operators in CABL}}
\label{sssec:edge}
Ablation experiments in Table.\ref{tab:a5} compared CABL module performance using Prewitt vs. Sobel edge detection operators. The Sobel operator achieved a 0.002 higher average I-AUC and 0.029 higher P-F1 than Prewitt. This is attributed to Sobel’s weighted convolution kernel, which enhances central pixel sensitivity and noise resistance compared to Prewitt’s uniform weights (as analyzed in Sec.\ref{ssec:CABL}). Based on these findings, we adopted the Sobel operator for our CABL module.

\subsubsection{\textbf{Ablation Study on Different Prompts in CGSR}}

Ablation experiments (Table.\ref{tab:a4}) show that combining bounding box and point prompts (with positive/negative point constraints) achieves the best segmentation performance.  The CGSR module effectively improves the segmentation accuracy of coarse-grained mask localization.

\section{Conclusion}

Our proposed framework integrates CABL and CGSR modules for weakly supervised image manipulation localization, achieving state-of-the-art performance in detection and segmentation using only image-level binary labels. The model demonstrates robust generalization across diverse datasets. This work highlights the efficacy of weakly supervised methods for high-fidelity manipulation localization and provides a strong foundation for future advancements in this domain.

\bibliographystyle{IEEEbib}
\bibliography{main}

\end{document}